\documentclass{article}

\usepackage{arxiv}
\usepackage{amsmath,amssymb,amsfonts}
\usepackage{float}
\usepackage{placeins}
\usepackage{soul}
\usepackage{algorithm}
\usepackage{graphicx}
\usepackage{textcomp}
\usepackage{comment}
\usepackage{booktabs}
\usepackage{pifont}

\usepackage{subfiles}
\usepackage[table,xcdraw]{xcolor}
\usepackage{multirow}
\usepackage{array}

\usepackage{listings}
\usepackage{ifthen}
\usepackage{fancyhdr}
\usepackage{soul}
	\makeatletter \def\@fancyvbox#1#2{\vbox{#2}} \makeatother
\usepackage[font=small,format=plain,labelfont= sc]{caption}
\usepackage[utf8]{inputenc}
\usepackage[T1]{fontenc}
\usepackage{color}
\usepackage{hyperref}
\usepackage{booktabs,multirow}
\usepackage{microtype}
\usepackage{tikz,bibunits,graphicx,circuitikz}
\usepackage{paralist}

\usepackage{pgfplots, pgfplotstable}
\usepgfplotslibrary{statistics}
\usetikzlibrary{patterns,intersections}
\pgfplotsset{compat=newest}
\usepackage{makecell}
\usepackage{tablefootnote}
\usepackage{float,placeins}
\usepackage{cellspace}
\usepackage{longtable}
\usepackage{xspace}
\usepackage{colortbl}
\usepackage{makecell} 

\newcolumntype{M}[1]{>{\centering\arraybackslash}m{#1}} 

\def\BibTeX{{\rm B\kern-.05em{\sc i\kern-.025em b}\kern-.08em
T\kern-.1667em\lower.7ex\hbox{E}\kern-.125emX}}

\newcommand{\red}[1]{\textcolor{red}{#1}}
\newcommand{\blue}[1]{\textcolor{blue}{#1}}

\hypersetup{
    colorlinks=true,
    linkcolor=black,
    filecolor=magenta,      
    urlcolor=blue,
    citecolor=blue,
}

\lstdefinestyle{busquedas}{
    basicstyle=\ttfamily\footnotesize,
    breakatwhitespace=false,         
    breaklines=true,                 
    captionpos=t,                    
    keepspaces=true,                
    showspaces=false,                
    showstringspaces=false,
    showtabs=false
}

\newcolumntype{?}{!{\vrule width 2pt}}

\begin{document}

\title{A XAI-based Framework for Frequency Subband Characterization of Cough Spectrograms in Chronic Respiratory Disease}

\author{%
\textbf{Patricia Amado-Caballero}$^{1}$\thanks{Email: patricia.amado@uva.es} \and
\textbf{Luis M. San-José-Revuelta}$^{1,2}$\thanks{Email: lsanjose@tel.uva.es} \and
\textbf{Xinheng Wang}$^{3,4}$\thanks{Email: H.Wang@ciptechnology.co.uk, xinheng.wang@xjtlu.edu.cn} \and
\textbf{José Ramón Garmendia-Leiza}$^{5}$\thanks{Email: jgarmendia@saludcastillayleon.es} \and
\textbf{Carlos Alberola-López}$^{1}$\thanks{Email: caralb@tel.uva.es} \and
\textbf{Pablo Casaseca-de-la-Higuera}$^{1,*}$\thanks{*Corresponding author. Email: casaseca@lpi.tel.uva.es}%
}

\date{
  $^{1}$ Laboratorio de Procesado de Imagen. ETSI de Telecomunicación, Universidad de Valladolid, Valladolid, Spain \\
  $^{2}$ Instituto de Neurociencias de Castilla y León (INCYL), Universidad de Salamanca, Salamanca, Spain \\
  $^{3}$ CIP Technologies LTD, Edinburgh, United Kingdom \\
  $^{4}$ School of Advanced Technology, Xi’an Jiaotong-Liverpool University, Suzhou 215123, China \\
  $^{5}$ Gerencia Regional de Salud de Castilla y León, Palencia, Spain
}

\maketitle

\begin{abstract}
This paper presents an explainable artificial intelligence (XAI)-based framework for the spectral analysis of cough sounds associated with chronic respiratory diseases, with a particular focus on Chronic Obstructive Pulmonary Disease (COPD). A Convolutional Neural Network (CNN) is trained on time–frequency representations of cough signals, and occlusion maps are used to identify diagnostically relevant regions within the spectrograms. These highlighted areas are subsequently decomposed into five frequency subbands, enabling targeted spectral feature extraction and analysis. The results reveal that spectral patterns differ across subbands and disease groups, uncovering complementary and compensatory trends across the frequency spectrum. Noteworthy, the approach distinguishes COPD from other respiratory conditions, and chronic from non-chronic patient groups, based on interpretable spectral markers. These findings provide insight into the underlying pathophysiological characteristics of cough acoustics and demonstrate the value of frequency-resolved, XAI-enhanced analysis for biomedical signal interpretation and translational respiratory disease diagnostics.
\end{abstract}

\keywords{Deep Learning, eXplainable Artificial Intelligence, cough analysis, respiratory diseases, occlusion maps, spectral features.}

\section{Introduction}
\label{intro}

Respiratory diseases represent a significant global health challenge, contributing to high mortality rates and long-term disability~\cite{who17}. About 65 million people suffer from Chronic Obstructive Pulmonary Disease (COPD) and 3 million die from it each year, making it the third leading cause of death worldwide. Asthma is the most common chronic disease in childhood, affecting 14\% of children worldwide. Lung cancer kills 1.6 million people a year, making it the most lethal cancer. The pandemic unleashed by the outbreak of the SARS-COV-2 coronavirus in 2020, which causes COVID-19, resulted in more than 5 million deaths associated with respiratory failure~\cite{who21}. Chronic diseases such as COPD and cancer frequently degenerate into disability and dependency~\cite{ATS2005homecare}, requiring constant patient monitoring to control disease exacerbations and respiratory support needs. In the specific case of these chronic diseases, it has been shown that the damage caused by COVID-19 infection considerably increases the risk of critical hospitalisation and death~\cite{who21}. 

Continuous monitoring of these patients is therefore essential, not only to track their condition but also to enable early detection of respiratory disease exacerbations.
 
The 2018 European Commission study emphasized the importance of telemedicine in the management of chronic respiratory diseases~\cite{CE18}, highlighting the need for further research. Despite growing interest since the early 2010s~\cite{Audit11}, progress has been hindered by the subjective nature of symptom assessment. However, objective measurements and real-time monitoring are essential for early diagnosis and improved patient prognosis~\cite{Pinnock13}. Accordingly, previous research has explored the automatic detection and classification of cough events using traditional machine learning (ML) and deep learning (DL) techniques, the latter outperforming the former. Although these methods have shown promise, disease identification remains a challenge without large and diverse cough databases. Additionally, the ``black box'' nature of DL models limits understanding of the underlying disease mechanisms.

This paper employs eXplainable Artificial Intelligence (XAI) to extract band-specific spectral features to identify cough audio patterns related to chronic respiratory diseases in general and COPD in particular. Occlusion maps~\cite{zeiler2014visualizing} are generated from a Convolutional Neural Network (CNN) to identify and highlight relevant regions in cough spectrograms. These map-weighted spectrograms are then divided into five distinct frequency bands for the extraction of targeted spectral features. Our findings reveal that band-specific features offer greater discriminatory power than those extracted from the entire weighted spectrogram. This approach uncovers significant differences not only between COPD and other disease groups but also between chronic and non-chronic patients, thereby deepening the understanding of disease-specific acoustic patterns and increasing the potential accuracy of diagnosis.

The structure of the paper is as follows: Section \ref{sota} reviews the state of the art. Section \ref{matmet} describes the proposed methodology along with the materials used for its evaluation. The main results are presented in Section \ref{results}. Finally, Section~\ref{sec:Disc} discusses these results and summarizes the key insights and conclusions of the study.

\section{State of The Art}
\label{sota}
In this section, we summarize the work to date investigating the automatic detection and/or classification of cough audio recordings, especially those related to chronic diseases and COPD. We begin by detailing the approaches used for cough detection, both traditional (not based on deep learning) and based on it. We then summarize the work focused on discriminating respiratory pathologies, and finally, we focus on studies that have attempted to characterize cough signals in COPD, especially those that seek to provide interpretability to processing systems.

\subsection{Automatic cough detection}

Traditional machine learning (ML) approaches for automatic cough detection \cite{drugman2013objective,Birring2008,Vizel2010,sterling2014automated,drugman2014using,amrulloh2015automatic,monge2018robust,hoyos2017efficient,hoyos2018efficient,monge2018audio,Monge19} typically rely on pattern recognition algorithms that analyze handcrafted features extracted from the temporal and spectral domains of audio signals. More recently, deep learning (DL) methods have demonstrated superior performance, surpassing traditional techniques. Convolutional and recurrent neural networks (CNNs/RNNs) are commonly used to classify pre-computed feature sets such as Mel Frequency Cepstral Coefficients (MFCCs) \cite{tokuda94_icslp} or to learn discriminative representations directly from raw audio data \cite{liu2014cough,amoh2015deepcough,amoh2016deep,kadambi2018,kvapilova2020continuous,li2020eeg,you2022automatic} for cough identification.

\subsection{Discriminatory analysis of pathologies}

Current research is focused on discrimination between different respiratory pathologies by characterizing the cough signals associated with each. For example, in the pediatric field, Swarnkar~\cite{swarnkar2013automatic} and Habashy~\cite{habashy2022cough} discriminated between dry and wet coughs, the latter using Audio Spectrogram Transformers (AST) achieved an 81\% F1 score. In 2013, Abeyratne et al. used support vector machines (SVM) to diagnose pneumonia in adults~\cite{abeyratne2013cough}. Later, in 2018, Sharan et al.~\cite{sharan2018automatic} used a similar approach to diagnose croup (an upper respiratory tract disease that often affects infants and children).

Traditional neural networks were also used to discriminate between pneumonia and asthma by Amrulloh et al. in ~\cite{amrulloh2015cough}. In 2022, Kumar et al. proposed a DL-based model for the classification of 10 common lung diseases in Indian adolescents~\cite{Kumar2022}. Interested readers are referred to~\cite{Ijaz2022} for a compendium of various works in the field of AI for the diagnosis of different lung diseases.

Due to their quantity and importance in the last five years, work related to cough-based COVID-19 detection deserves specific mention. A review of works published up to 2022 including several DL-based methods can be found in~\cite{Ghrabli22}. Details on some relevant contributions follow: in 2020, Laguarta et al.~\cite{laguarta2020covid} achieved a sensitivity of 98.5\% using CNNs with MFCC inputs, while Imran et al.~\cite{imran2020ai4covid} combined CNNs with SVMs, reaching 92.64\% accuracy. Tena et al.~\cite{tena2022automated} used the deep network YAMNET to obtain almost 90\% accuracy. Pahar et al. \cite{pahar2020covid} used both a ResNet50 model (a type of convolutional neural network, CNN) and a Long Short-Term Memory (LSTM) network (a type of recurrent neural network, RNN, that employs memory cells and gating mechanisms to capture long-range temporal dependencies), achieving AUCs of 98\% and 94\%, respectively. In 2021, Vrindavanam et al. \cite{vrindavanam2021machine} achieved an accuracy of 70.6\% with a CNN, and Feng et al. \cite{Feng2021DeeplearningBA} obtained 90\%  using an RNN on the Coswara dataset \cite{bhattacharya2023coswara}, dropping to 80\% when combined with the Virufy~\cite{Kha21} dataset.
More recently, Hussain et al.~\cite{hussain2024cough2covid19} used a multilayer ensemble deep learning (MLEDL) model that incorporates MFCC, spectrograms, and chromagrams, reaching 98\% accuracy across public datasets. Similarly, Islam et al.~\cite{islam2025robustcovid19detectioncough} used deep neural decision trees and random forests, reporting an accuracy up to 97\% across diverse datasets.

\subsection{Characterization of cough audio in COPD and other respiratory diseases. Application of Explainable Artificial Intelligence (XAI)}

In this section, we summarize the methods that have focused on cough analysis in chronic diseases and, in particular, COPD. Most recent methods employ deep machine learning tools to solve the drawbacks of their initial versions based mainly on temporal, frequency, or mixed hand-crafted features extracted from cough audios. Their main disadvantage is the lack of interpretability of the networks that perform the processing and optimization. Therefore, in subsection \ref{sec_xai}, we focus on the few previous references that have attempted to employ eXplainable AI (XAI), which is where our proposal fits in.

\subsubsection{COPD detection and charactization}
Recent studies demonstrate the potential of AI and deep learning to analyze cough sounds related to COPD and other chronic respiratory conditions. For example, Sharan et al. \cite{Sharan2018} used cough recordings from an iPhone to estimate spirometric parameters in patients with COPD, finding a strong correlation with measures like FEV1 and FVC. Later, Srivastava et al. \cite{Srivastava2021} applied CNNs with features such as MFCCs and Mel spectrograms to detect COPD, achieving up to 93\% accuracy. Balamurali et al. \cite{Balamurali2021} used a BiLSTM model to classify children’s coughs, with accuracies over 84\% for healthy vs. pathological and 91\% for specific conditions, though performance dropped for multi-class tasks. More recently, Basha and Reddy \cite{Basha2023} proposed an end-to-end framework combining segmentation, feature extraction, and classification for various lung diseases, reporting moderate precision for distinguishing disease subtypes. Additionally, Sánchez-Morillo et al. \cite{SanchezMorillo2024} highlighted the challenge of discriminating cough sounds from background noise in their systematic review. All of these studies underscore the potential for using artificial intelligence and DL techniques in cough sound analysis related to COPD, although they also highlight the challenges associated with accuracy and differentiation between various pathologies. However, though DL consistently outperforms traditional methods, its “black box” nature remains a limitation. Integrating eXplainable AI (XAI) could help reveal disease-specific cough patterns and improve diagnosis and follow-up of chronic diseases.

\subsubsection{XAI-informed DL methods for cough characterization}
\label{sec_xai}

Several works have employed XAI-informed DL methods for cough characterization. Wullenweber et al.~\cite{wullenweber2022coughlime} employed Local Interpretable Model-Agnostic Explanations (LIME)~\cite{ribeiro2016should} to identify the relevance of different audio-cough components in the diagnosis of COVID-19. Audio-cough spectrograms were decomposed into interpretable components which were further analyzed for importance in the output of a deep ResNet model. Shen et al.~\cite{Shen2024} used Gradient-weighted Class Activation Mapping (Grad-CAM)~\cite{selvaraju2017grad} to visualize the contribution of CNN-extracted features for differentiating COVID-19, asthma, pneumonia, and healthy subjects. Some other recent studies have explored hybrid and transformer-based models for the detection of COVID-19 through cough sounds~\cite{avila2021investigating,sobahi2022explainable}. In~\cite{avila2021investigating}, feature engineering techniques are combined with a CNN model that explains itself using the score-class activation map (CAM) technique, achieving an AUC of 0.81, while~\cite{sobahi2022explainable} employs the YAMNet model and fractal transformation techniques to classify cough sounds into categories, reaching accuracies above 98\% on multiple datasets. The authors also demonstrate the effectiveness of attention mechanisms in the Vision Transformer (ViT) for cough sound-based COVID-19 detection, providing a visual explanation of the ViT model’s inner workings. 

More related with the present proposal, in \cite{amado2024}, occlusion maps were employed to highlight spectrogram regions most relevant to CNN-based cough detection. A Gaussian Mixture Model (GMM) fitted to these weighted spectrograms revealed disease-related energy variations, but suffered from two limitations: (1) the bimodal GMM often failed to accurately model certain spectrograms, leading to potential misinterpretations, and (2) the explainability relied mainly on temporal differences, making the method highly sensitive to the precise onset and offset of cough events. These issues were addressed in \cite{amado2025}, which introduced a spectral analysis of the map-weighted spectrograms, to identify key spectral information yielding a more discriminative approach.

\section{Materials and methods}
\label{matmet}

\subsection{Materials}
\label{mat}
For this study, full day prospective audio recording was performed over ambulatory patients with persistent cough under real living conditions. The database comprises 17 adult patients aged between 23-87 with different respiratory pathologies. These patients were divided into six study groups, as detailed in Table \ref{tab1:patients}. Audio samples were collected using a Sony Xperia 72 Android Smarthphone with a sampling rate of 44.1 kHz. 

\begin{table*}[htbp]
\centering
\caption{Summary of the patient database}
\resizebox{\textwidth}{!} {
\begin{tabular}{|c|M{3cm}|M{2cm}|M{4cm}|M{2cm}|M{6cm}|}
\hline
\textbf{GROUP} & \textbf{Comparison}& \textbf{Patients (C1)} & \textbf{Pathologies (C1)}&\textbf{Patients (C2)} & \textbf{Pathologies (C2)}   \\
\hline
\noalign {\hrule height 1pt }
\hline  \hline
\textbf{G1}& Chronic (C1) vs. Non-Chronic (C2)                  & 11                      & COPD (6), asthma (1), bronchiectasis (1), sarcoidosis (1), lung cancer (2) & 6                       & Acute respiratory disease (ARD, 3), pneumonia (3)  \\
\hline
\textbf{G2}  & COPD (C1) vs. Other diseases (C2)                  & 6                       & COPD                                                                       & 11                      & Acute respiratory disease (ARD, 3), pneumonia (3),  asthma (1), bronchiectasis (1), sarcoidosis (1), lung cancer (2) \\ \hline
\textbf{G3}  & COPD (C1) vs. Other diseases excluding cancer (C2) & 6                       & COPD                                                                       & 9                       & Acute respiratory disease (ARD, 3), pneumonia (3),  asthma (1), bronchiectasis (1), sarcoidosis (1)               \\
\hline
\textbf{G4}&COPD (C1) vs. ARD and pneumonia (C2)               & 6                       & COPD                                                                       & 6                       & Acute respiratory disease (ARD, 3), pneumonia (3)   \\
\hline
\textbf{G5}  & COPD (C1) vs. Other chronic diseases (C2)          & 6                       & COPD                                                                       & 5                       & Asthma (1), bronchiectasis (1), sarcoidosis (1), lung cancer (2)     \\
\hline
\textbf{G6} & COPD (C1) vs. Lung cancer (C2)                     & 6                       & COPD                                                                       & 2                       & Lung Cancer       \\
\hline
\end{tabular}

}
\label{tab1:patients}
\end{table*}

\subsection{Methods}
\label{sec:Methods}

In this paper, we apply Explainable Artificial Intelligence (XAI) techniques to extract frequency band–specific characteristics to identify audio patterns associated with chronic respiratory diseases in general, and COPD in particular. Similar to \cite{amado2024,amado2025}, we employ occlusion maps in a CNN model to identify critical regions within cough spectrograms that contribute most to cough identification. Compared to other perturbation-based methods such as LIME or SHAP, occlusion maps are computationally simpler and directly tied to the model’s prediction on the original input. Unlike gradient-based methods such as Grad-CAM or attention mechanisms, which may highlight broad or diffuse regions that do not always correspond to truly discriminative features, occlusion maps yield more localized and interpretable explanations. This makes them especially suited for complex signals like cough spectrograms where localized energy patterns carry diagnostic information. 

\subsubsection{Audio signal preprocessing and cough identification}

We use spectrograms ---2D representations of audio in the time-frequency domain--- as input to a CNN that distinguishes cough sounds from other audio signals.
To obtain these spectrograms, the original audio signals were subsampled ($5\times$ down-sampling) to a frequency of 8.82 kHz. After that, the power spectral density (PSD) was computed for $10$ ms segments using a Hanning window with no overlap. We concatenated these segments into 1-second intervals, keeping only the positive frequencies due to conjugate symmetry, and we log-normalized them, obtaining the final $45\times100$ spectrograms for the CNN.

The CNN employed for cough identification was designed from scratch. The architecture consists of a first convolutional layer with 32 filters, using 2x2 kernels and ReLU activation, followed by a 2x2 Max-Pool layer to reduce dimensionality. Then, a Dropout layer (with an initial rate of $10\%$ to avoid overfitting) follows. This structure (Convolutional + Max-Pool + Dropout) is repeated, but this time with twice as many filters (64 filters). Before the output layers, there is an additional stage: a convolutional layer with 128 filters, another Dropout layer, a convolutional layer with 256 filters, and a final Max-Pool layer. The output of this sequence is resized ("flattened") to feed the Fully-Connected layers. The first Fully-Connected layer has 512 neurons and ReLU activation. The second Fully-Connected layer has two neurons and uses a softmax activation function for the final binary classification (cough/non-cough).

The training process was carried out using the AdaMax optimizer ($\alpha=0.002$), batch size$=128$ and 50 epochs. To avoid overfitting, we employed a validation set comprising $20\%$ of the training dataset, which itself constituted $80\%$ of the entire audio clips collection. Train and test sets were organized in 5-folds to ensure that coughs belonging to patients in the training set were never in its corresponding test set in the fold and all the patients were included at least once in a test set.  

\begin{figure}[!htb]
\centering
\includegraphics[width=0.8\textwidth]{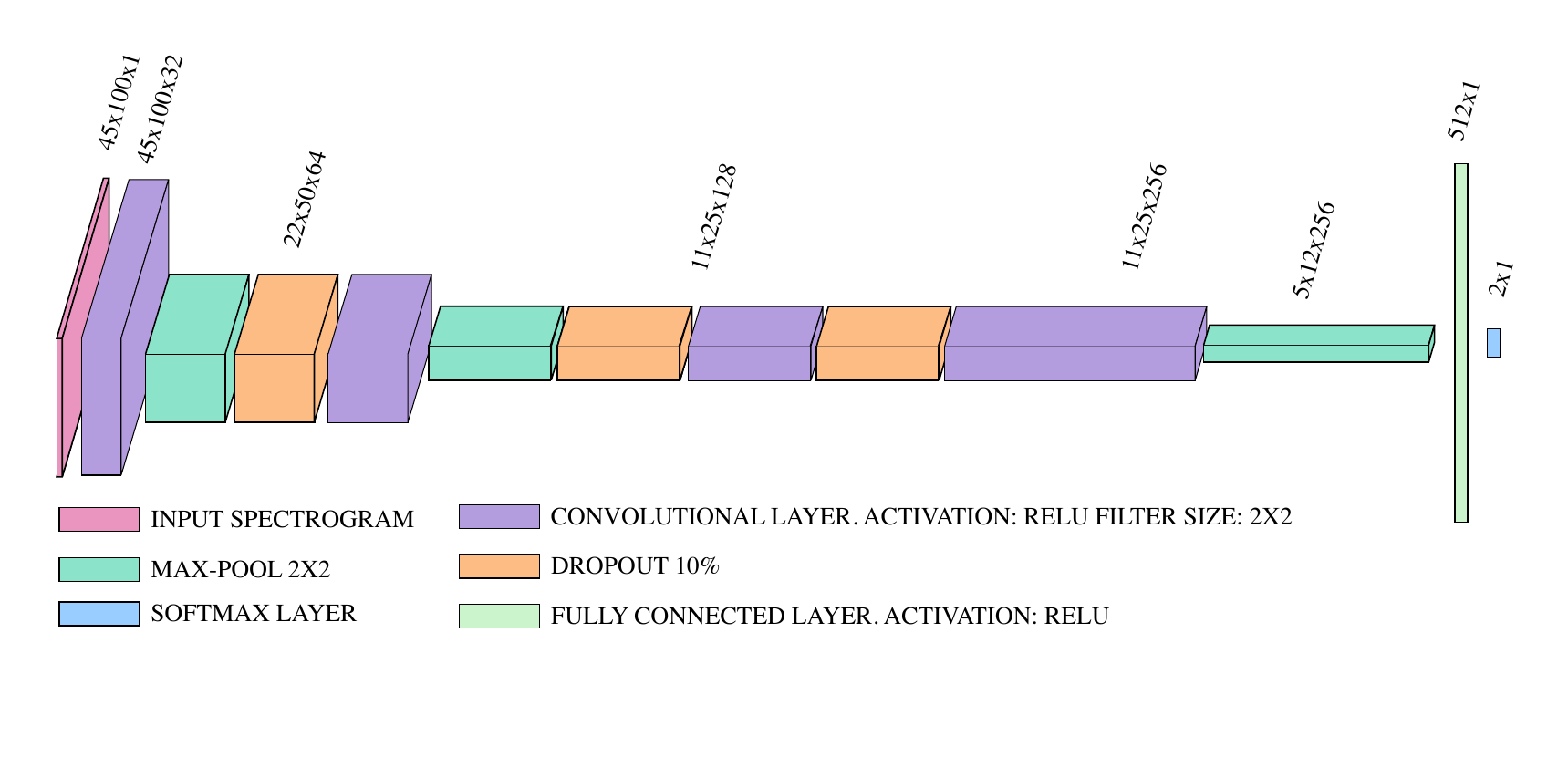}
\captionsetup{justification=centering}
\caption{CNN architecture for cough detection.}
\label{figCNN}
\end{figure}

\subsubsection{XAI-driven cough analysis} 
\begin{figure*}[!htb]
\centering
\includegraphics[width=1\textwidth]{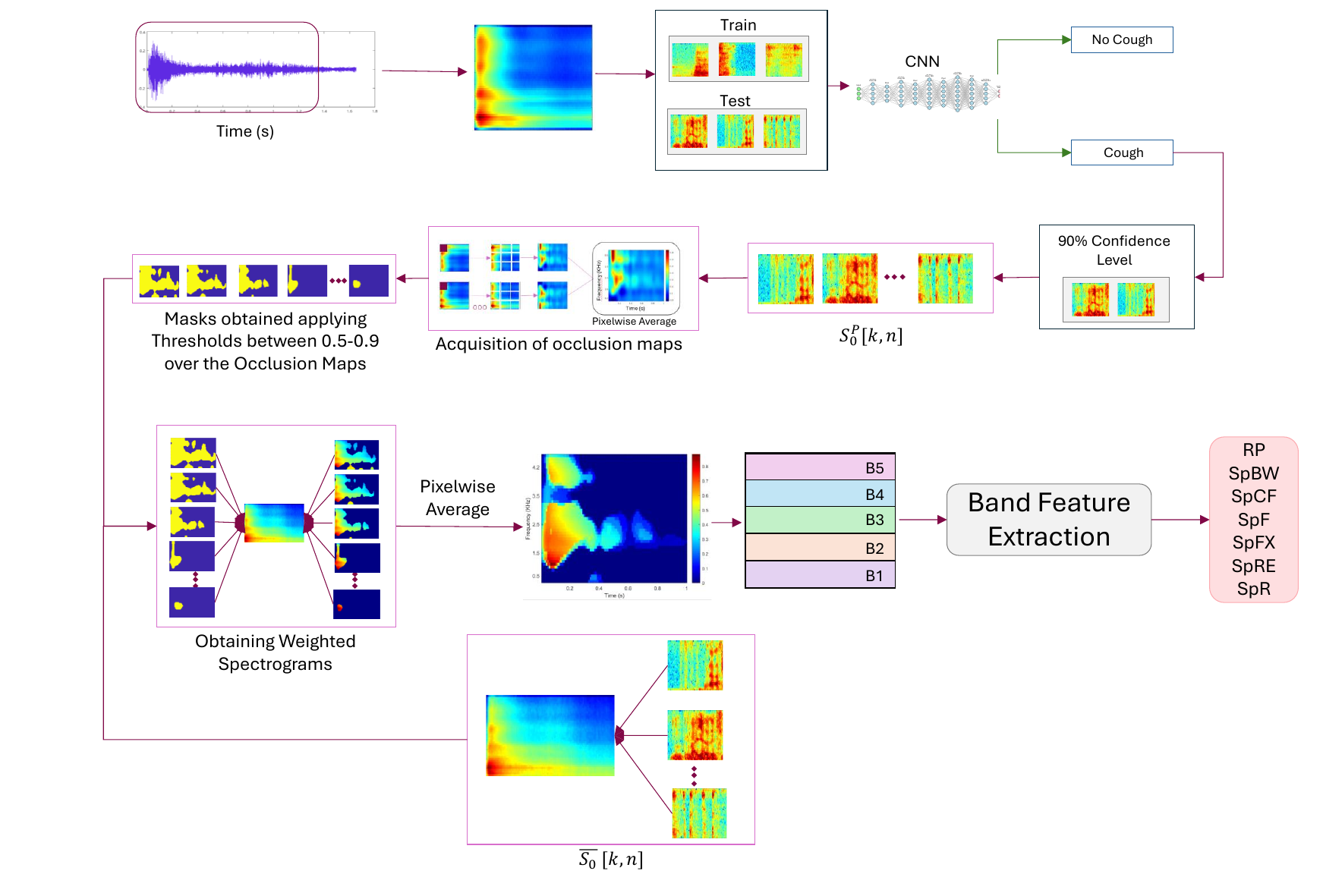}
\captionsetup{justification=centering}
\caption{Overview of the classification and XAI-driven cough analysis.}
\label{figMet}
\end{figure*}

After training the CNN, we sought to interpret its decisions and identify which features of the cough sound were relevant to the model. Once these features are identified, we quantitatively analyze them to find differences between the groups. This is explained in detail in section~\ref{subsec:Features}. Prior to this, we use XAI techniques to highlight significant regions of the spectrograms, as follows: 

\begin{itemize}
    \item Those spectrograms classified as cough with a high confidence level ($\geq90\%$) in the Softmax output, were selected. Let $\mathcal{S}^p_0[k,n]$ denote the $p$-th of those spectrograms, $0\leq p \leq P-1$. For each of these  spectrograms, we computed the occlusion map, a XAI technique consisting of applying a sliding mask over the spectrogram, temporarily hiding part of its information from the CNN. The class probability obtained with the modified input allows us to estimate the importance of the hidden area for the CNN decision (a higher probability indicates lower importance of the hidden area). This process is repeated across the entire spectrogram, and the values are stored in a matrix of the original size of the spectrogram. Let $\mathcal{M}^p[k,n]$ denote the occlusion map associated to $\mathcal{S}^p_0[k,n]$.  Then, a representative occlusion map of each patient's cough is obtained by normalizing all individual occlusion maps and pixelwise averaging, i.e., 
    \begin{align}
    \overline{\mathcal{M}}[k,n] &= \frac{1}{P} \sum_{p=0}^{P-1}  \mathcal{M}^p[k,n].
\end{align}
    This averaged map highlights the spectral regions (time-frequency) that the CNN considered most important for detecting that patient's cough. We then find the $\text{Th}$-percentile across the values of $\overline{\mathcal{M}}[k,n]$, i.e., 
    \begin{align}
    \alpha &= 
\mathop{\mathrm{\text{percentile}}}\limits_{k,n} \left (\overline{\mathcal{M}}[k,n],\text{Th} \right ) 
\label{eq:Th}
\end{align}    
    \item Once we obtained the occlusion map, each patient’s pixel-averaged spectrogram was weighted by its thresholded occlusion map, i.e., 
\begin{align}
    \overline{\mathcal{S}}_0[k,n] &= \frac{1}{P} \sum_{p=0}^{P-1}  \mathcal{S}_0^p[k,n] \\
    \mathcal{S}[k,n] &= \overline{\mathcal{S}}_0[k,n] \odot  
    \left [\overline{\mathcal{M}}[k,n] > \alpha \right ]
\end{align}
where $\odot$ denotes the Hadamard  product and the operation within brackets yields a binary mask resulting from the comparison between $\overline{\mathcal{M}}[k,n]$ and $\alpha$. 

The result is a masked spectrogram that amplifies the spectral relevant areas for cough detection, improving the interpretability of the features. We refer to this as ``Weighted Spectrogram'' as a general term, although in this specific case, the weighting is performed through a binary mask. The overall process is illustrated in Figure \ref{figMet}.
\end{itemize}

\subsubsection{Band-specific Spectral features obtained from the Weighted Spectrograms}  
\label{subsec:Features}

Spectral features~\cite{ramalingman2006} have been used to characterize disease-specific cough signatures~\cite{amado2025}, offering a physically meaningful representation without requiring precise cough onset detection~\cite{amado2024}. Unlike~\cite{amado2025}, which extracted features across the entire $[0, 4.41)$ kHz range, we segmented map-highlighted regions into five frequency bands to capture more granular, frequency-specific features. This band-specific approach, inspired by~\cite{Monge19}, enhances discriminatory power and allows the identification of disease-associated patterns from weighted spectrograms. By focusing on sub-band features, we reveal significant inter-disease differences that can be hindered when performing a whole-spectrum analyses.

The calculated features are described in the following paragraphs. We refer to the discrete (weighted) one-sided spectrogram as $\mathcal{S}[k,n]$, where $k=0,\ldots,K$ denotes de frequency index ($K=44$), being the discrete frequencies $f[k]=k\cdot f_s/(2\cdot K+1)$, with $f_s=44.1/5=8.82$ kHz. On the other hand, $n=0,\ldots,N-1$ stands for the time index ($N=100$). 

Five frequency sub-bands $B_j, j=1,\ldots,5$ have been considered, and subband-specific weighted spectrograms have been defined as follows: 

\begin{equation}
\mathcal{S}_j[k,n]=\left\{\begin{array}{cl}
\mathcal{S}[k,n]&; f[k]\in B_j\\
0&; \text{elsewere}
\end{array}
\right.
\label{eq:Bj}
\end{equation}
with $B_1\equiv[0, 0.5)$ kHz, $B_2\equiv[0.5, 1.0)$ kHz, $B_3\equiv[1.0, 1.5)$ kHz, $B_4\equiv[1.5, 2.0)$ kHz, $B_5\equiv[2.0, 4.41)$ kHz. To compare with the approach in~\cite{amado2025}, we also calculated the features for the whole band $B\equiv[0, 4.41)$ kHz. In that case, $\mathcal{S}[k,n]$ replaces $\mathcal{S}_j[k,n]$ in the feature equations.

The first extracted feature is the sub-band \textbf{Relative power}, obtained as the ratio between the power in the $j-th$ sub-band and the total power:  
\begin{equation}
RP_j= \frac{1}{N}\sum_{n=0}^{N-1}\left[\frac{\sum\limits_{k=0}^{K}\mathcal{S}_j[k,n]}{\sum\limits_{k=0}^{K}\mathcal{S}[k,n]}\right]   \label{eq:RP}
\end{equation}

For the whole band $B$, this feature was obtained as the relative AC power, since the equivalent $RP$ would be equal to one in all cases:

\begin{equation}
AC=RP_B= \frac{1}{N}\sum_{n=0}^{N-1}\left[\frac{\sum\limits_{k=1}^{K}\mathcal{S}[k,n]}{\sum\limits_{k=0}^{K}\mathcal{S}[k,n]}\right]   
\label{eq:AC}
\end{equation}

The band-specific \textbf{Spectral bandwidth} was calculated as a measure of the spread of the spectral distribution as
\begin{equation}
SpBW_j=\frac{1}{N}\displaystyle\sum\limits_{n=0}^{N-1}\left[\frac{\sum\limits_{k=0}^{K}(f[k]-SpC_j[n])^2\cdot\mathcal{S}_j[k,n]}{\sum\limits_{k=0}^{K}\mathcal{S}_j[k,n]}\right]
\label{eq:Specband}
\end{equation}
where $SpC_j[n]$ is the band-specific \textbf{Spectral centroid}, representing the sub-band spectral center of gravity as:
\begin{equation}
SpC_j[n]= \frac{\sum\limits_{k=0}^{K}f[k]\cdot\mathcal{S}_j[k,n]}{\sum\limits_{k=0}^{K}\mathcal{S}_j[k,n]}\label{eq:SpecCent}
\end{equation}

The band-specific \textbf{Spectral Crest Factor} was also computed as a feature used to detect the dominant frequency in the sub-band. 

\begin{equation}
SpCF_j= \frac{1}{N}\displaystyle\sum_{n=0}^{N-1}\left[
\frac{\max\limits_k(\mathcal{S}_j[k,n])}{C\cdot\sum\limits_{k=0}^{K}\mathcal{S}_j[k,n]}
\right]
\label{eq:SpecCrest}
\end{equation}
with $C= 1/(\max(f[k])-\min(f[k])+1), f[k]\in B_j$.

The band-specific \textbf{Spectral flatness} was obtained as a measure quantifying the degree of flatness of the spectrum in each sub-band. 
\begin{equation}
SpF_j=\frac{1}{N}\displaystyle\sum\limits_{n=0}^{N-1}\left[
\frac{\exp\left(\mathbb{E}\left\{\log(\mathcal{S}_j[k,n])\right\}\right)}{\mathbb{E}\left\{\mathcal{S}_j[k,n]\right\}} 
\right]
\label{eq:SpecFlat}
\end{equation}
where $\mathbb{E}\{\cdot\}$ denotes the expectation operator, evaluated by averaging over the frequency index.

To quantify the spectral variation between two consecutive time stamps for each sub-band, we computed the band-specific \textbf{Spectral flux}~\cite{giannakopoulos2014introduction}:
\begin{equation}
SpFX_j= \frac{1}{N-1}\displaystyle\sum_{n=1}^{N-1}\left[
\sum_{k=0}^{K}(\mathcal{S}_j[k,n]-\mathcal{S}_j[k,n-1])
\right]
\label{eq:SpecFlux}
\end{equation}

The band-specific \textbf{Spectral Renyi entropy} was also computed as a generalized measure of uncertainty or randomness in each sub-band.
\begin{equation}
SpRE_j= \frac{1}{N}\displaystyle\sum_{n=0}^{N-1}\left[
\frac{1}{1-q}\cdot\log\left(\sum_{k=0}^{K}{\mathcal{S}_j[k,n]}\right) ^q 
\right]
\label{eq:SpecRenyiEnt}
\end{equation}
where $q=4$ was used.

Finally, we defined the band-specific \textbf{Spectral roll-off} to account for the frequencies in each sub-band representing the 85th percentile of the total power~\cite{giannakopoulos2014introduction}:
\begin{equation}
    SpR_j=\frac{1}{N}\displaystyle\sum_{n=0}^{N-1}f[k^j_{85}[n]]
 \label{eq:SpecRollOff}
\end{equation}
with $k^j_{85}[n]$ the minimum value of $k$ for which:
\begin{equation}
\sum_{k=0}^{k^j_{85}[n]}\mathcal{S}_j[n,k] \geq 0.85\cdot\sum_{k=0}^{K}\mathcal{S}_j[n,k]
\end{equation}

\subsubsection{Testing for group differences}

To identify statistically significant differences between groups in each set (G1--G6) defined in section \ref{mat}, the spectral features extracted for each group were analyzed using hypothesis testing. Initially, a Gaussianity test was performed. If both groups satisfied Gaussianity, an unpaired Student’s t-test was applied. Otherwise, the Mann-Whitney U-test was used. After conducting the tests for all features, boxplots were generated for cases where significant differences were identified (p-value < 0.05).

\section{Results}
\label{results}

Tables \ref{tab:features_resume1} and \ref{tab:features_resume2} provide an overview of the results obtained in the study. Specifically, table  \ref{tab:features_resume1} indicates which groups show significant differences for each feature across frequency bands and table~\ref{tab:features_resume2} identifies a summary of the bands over the groups.
Tables \ref{tab:features_band1}---\ref{tab:features_band5} present separability results for groups G1---G6 for the five proposed frequency bands. The $p$-values reported in the table show the best obtained value for all evaluated thresholds ---see Equation \eqref{eq:Th}--- over the weighted spectrograms in each band. Statistically significant differences are obtained in each group for at least one band-specific feature. For the sake of comparison, table \ref{tab:bandaGlobal} shows the equivalent results for the whole $B\equiv[0-4.41)$ kHz band as in~\cite{amado2025}, where no significant differences can be observed for groups G1 or G6, even though notably low $p$-values are obtained for some feature-group combinations in groups G2--G5. Figures~\ref{fig:RPBox_1}--\ref{fig:SPR_3} present boxplots of all XAI-driven spectral features showing statistically significant differences in the comparison groups across any of the five frequency bands. For the sake of comparison, the equivalent boxplots for the entire band $B$~\cite{amado2025} are also included. The presented results are discussed in Section \ref{sec:Disc}.

\newcommand{\cmark}{\ding{51}}  
\newcommand{\xmark}{\ding{55}}  

\begin{table*}[!htb]
\caption{Summary of Study Groups Showing Statistically Significant Differences (p < 0.05) for Each Feature (rows) Across  Frequency Bands (columns). Red means significantly higher Feature values for C1 (see table~\ref{tab1:patients}) than C2. Blue, opposite meaning.} 
\label{tab:features_resume1}
\resizebox{\textwidth}{!} {
\begin{tabular}{@{}|c|c|c|c|c|c|c|@{}}
\toprule
\textbf{Feature} & \textbf{Band 1} & \textbf{Band 2} & \textbf{Band 3} & \textbf{Band 4} & \textbf{Band 5} &  \textbf{Global Band~\cite{amado2025}} \\ \midrule
{$RP$}      & \red{G1} \red{G2} \red{G3} \red{G4}     & \red{G1} \red{G2} \red{G4}        & \blue{G2} \blue{G3} \blue{G4}        & \blue{G1} \blue{G2} \blue{G3} \blue{G4}     & \red{G2} \red{G3} \red{G4}        & \blue{G2} \blue{G3} \blue{G4}         \\ \midrule
{$SpBW$}    & \red{G2}              & \red{G5}              & \red{G2 G3 G4 }      &\blue{G2 G3 G4}        & \red{G1} \blue{G4}           & \red{G2 G3 G4}         \\ \midrule
{$SpCF$}    & \blue{G2 G3}           & \blue{G1 G2 G3 G4}     & \blue{G2 G3 G4}        & \red{G2 G3 G4}        & \blue{G2 G3 G4}        & \red{G1} \blue{G2 G3 G4}      \\ \midrule
{$SpF$}     & \red{G2}              & \red{G1}              & \red{G2 G3 G4}        & \blue{G2 G3 G4}        &         & \red{G2 G3 G4 G5}      \\ \midrule
{$SpFx$}    & \blue{G2}              & \blue{G2 G5}           & \red{G2 G6}           & \blue{G2 G3 G4}        & \red{G2}              &                  \\ \midrule
{$SpRe$}    & \red{G2}             &                 & \red{G2 G3 G4}        & \red{G1 G2 G3 G4}     & \red{G1 G3 G4}        & \red{G2 G3 G4 G5}      \\ \midrule
{$SpR$}     &                 &                 &               & \red{G2 G3 G4 G5}     & \red{G1 G2 G3 G4}     & \red{G2 G3 G4 G5}      \\ \bottomrule
\end{tabular}
}
\end{table*}

\begin{table*}[!htb]
\caption{Summary of the Bands Showing Statistically Significant Differences (p < 0.05) for Each Group (rows) Across Features (columns). GB: Global Band (as in \cite{amado2025}. Red and blue colors have the samen meaning as in table~\ref{tab:features_resume1}.}
\label{tab:features_resume2}
\resizebox{\textwidth}{!} {
\begin{tabular}{cccccccl}
\toprule
\multicolumn{1}{l}{} & \multicolumn{7}{c}{\textbf{FEATURES}}                                                                                             \\ \midrule
\textbf{Study Group}       & \textbf{$RP$}    & \textbf{$SpBW$} & \textbf{$SpCF$}  & \textbf{$SpF$} & \textbf{$SpFx$} & \textbf{$SpRe$} & \multicolumn{1}{c}{\textbf{$SpR$}} \\ \midrule
\textbf{G1}          &  \red{1},\red{2},\blue{4} & \red{5} & \blue{2},\blue{GB}  & \red{2} && \red{4},\red{5} & \red{5}\\ \midrule
\textbf{G2}          & \red{1},\red{2},\blue{3},\blue{4},\red{5},\blue{GB}            & \red{1},\red{3},\blue{4},\red{GB} & \blue{1},\blue{2},\blue{3},\red{4},\blue{5},\blue{GB}     &  \red{1},\red{3},\blue{4},\red{GB}  & \blue{1},\blue{2},\red{3},\blue{4},\red{5}& \red{1},\red{3},\red{4},\red{GB}    &  \red{4},\red{5},\red{GB}                          \\ \midrule
\textbf{G3}          &  \red{1},\blue{3},\blue{4},\red{5},\blue{GB}  & \red{3},\blue{4},\red{GB} & \blue{1},\blue{2},\blue{3},\red{4},\blue{5},\blue{GB}      & \red{3},\blue{4}       & \blue{4}           &  \red{3},\red{4},\red{5},\red{GB}   & \red{4},\red{5},\red{GB}                         \\ \midrule
\textbf{G4}          & \red{1},\red{2},\blue{3},\blue{4},\red{5},\blue{GB}           &  \red{3},\blue{4},\blue{5},\red{GB}   & \blue{2},\blue{3},\red{4},\blue{5},\blue{GB}  & \red{3},\blue{4}      & \blue{4}            &   \red{3},\red{4},\red{5},\red{GB}   & \red{4},\red{5},\red{GB}                         \\ \midrule
\textbf{G5}          &                & \red{2}           &                &              & \blue{2}            &  \red{GB}       &\red{4},\red{GB}            \\ \midrule
\textbf{G6}          &                &               &                &              & \red{3}            &               &                                  \\ \bottomrule
\end{tabular}
}
\end{table*}

\begin{table*}[!htb]
\centering
\caption{Separability results obtained for groups G1--G6 in the 0--0.5 kHz frequency band with the proposed method. $p$-values are obtained from hypothesis testing over spectral features extracted from XAI-driven \textit{Weighted Spectrograms}. Statistical significance is considered for $p < 0.05$ and highlighted using green, boldfaced font.}
\label{tab:features_band1}
\begin{tabular}{lccccccc}
\toprule
 & \multicolumn{7}{c}{\textbf{Frequency Band 0--0.5 kHz}} \\
\midrule
 & \textbf{$RP_1$} & \textbf{$SpBW_1$} & \textbf{$SpCF_1$} & \textbf{$SpF_1$} & \textbf{$SpFx_1$} & \textbf{$SpRE_1$} & \textbf{$SpR_1$} \\
 & Th=0.7 & Th=0.5 & Th=0.5 & Th=0.5 & Th=0.5 & Th=0.5 & Th=0.5 \\
\midrule
\textbf{G1} & \textcolor{green}{\textbf{0.0273}} & 0.4609 & 0.3487 & 0.5233 & 0.5233 & 0.2998 & 0.9022 \\
\textbf{G2} & \textcolor{green}{\textbf{0.0202}} & \textcolor{green}{\textbf{0.0357}} & \textcolor{green}{\textbf{0.0357}} & \textcolor{green}{\textbf{0.0470}} & \textcolor{green}{\textbf{0.0470}} & \textcolor{green}{\textbf{0.0357}} & 0.0740 \\
\textbf{G3} & \textcolor{green}{\textbf{0.0256}} & 0.0655 & \textcolor{green}{\textbf{0.0480}} & 0.0859 & 0.0655 & 0.0655 & 0.1351 \\
\textbf{G4} & \textcolor{green}{\textbf{0.0042}} & 0.1277 & 0.0909 & 0.1277 & 0.1350 & 0.0909 & 0.2338 \\
\textbf{G5} & 0.4745 & 0.1667 & 0.1667 & 0.2381 & 0.0951 & 0.2381 & 0.2381 \\
\textbf{G6} & 0.1428 & 0.1429 & 0.2857 & 0.1429 & 0.1461 & 0.1429 & 0.1429 \\
\bottomrule
\end{tabular}
\end{table*}

\begin{table*}[!htb]
\centering
\caption{Separability results obtained for groups G1--G6 in the 0.5--1 kHz frequency band with the proposed method. $p$-values are obtained from hypothesis testing over spectral features extracted from XAI-driven \textit{Weighted Spectrograms}. Statistical significance is considered for $p < 0.05$ and highlighted using green, boldfaced font.}
\label{tab:features_band2}
\begin{tabular}{lccccccc}
\toprule
 & \multicolumn{7}{c}{\textbf{Frequency Band 0.5--1 kHz}} \\
\midrule
 & \textbf{$RP_2$} & \textbf{$SpBW_2$} & \textbf{$SpCF_2$} & \textbf{$SpF_2$} & \textbf{$SpFx_2$} & \textbf{$SpRE_2$} & \textbf{$SpR_2$} \\
 & Th=0.7 & Th=0.6 & Th=0.5 & Th=0.5 & Th=0.5 & Th=0.8 & Th=0.5 \\
\midrule
\textbf{G1} & \textcolor{green}{\textbf{0.0103}} & 0.7325 & \textcolor{green}{\textbf{0.0019}} & \textcolor{green}{\textbf{0.0273}} & 0.8075 & 0.3487 & 0.0997 \\
\textbf{G2} & \textcolor{green}{\textbf{0.0477}} & 0.3011 & \textcolor{green}{\textbf{0.0202}} & 0.5249 & \textcolor{green}{\textbf{0.0477}} & 0.0968 & 0.0747 \\
\textbf{G3} & 0.0663 & 0.0879 & \textcolor{green}{\textbf{0.0120}} & 0.6070 & 0.0663 & 0.1423 & 0.3360 \\
\textbf{G4} & \textcolor{green}{\textbf{0.0087}} & 0.3939 & \textcolor{green}{\textbf{0.0087}} & 0.2320 & 0.1168 & 0.1775 & 0.0606 \\
\textbf{G5} & 1.0000 & \textcolor{green}{\textbf{0.0238}} & 0.2619 & 0.2619 & \textcolor{green}{\textbf{0.0196}} & 0.3571 & 0.1667 \\
\textbf{G6} & 0.2857 & 0.2857 & 0.6429 & 0.6429 & 0.1037 & 0.2857 & 1.0000 \\
\bottomrule
\end{tabular}
\end{table*}

\begin{table*}[!htb]
\centering
\caption{Separability results obtained for groups G1--G6 in the 1--1.5 kHz frequency band with the proposed method. $p$-values are obtained from hypothesis testing over spectral features extracted from XAI-driven \textit{Weighted Spectrograms}. Statistical significance is considered for $p < 0.05$ and highlighted using green, boldfaced font.}
\label{tab:features_band3}
\begin{tabular}{lccccccc}
\toprule
 & \multicolumn{7}{c}{\textbf{Frequency Band 1--1.5 kHz}} \\
\midrule
 & \textbf{$RP_3$} & \textbf{$SpBW_3$} & \textbf{$SpCF_3$} & \textbf{$SpF_3$} & \textbf{$SpFx_3$} & \textbf{$SpRE_3$} & \textbf{$SpR_3$} \\
 & Th=0.7 & Th=0.7 & Th=0.5 & Th=0.7 & Th=0.5 & Th=0.7 & Th=0.5 \\
\midrule
\textbf{G1} & 0.0983 & 0.0782 & 0.1490 & 0.0782 & 0.0616 & 0.0983 & 0.0950 \\
\textbf{G2} & \textcolor{green}{\textbf{0.0145}} & \textcolor{green}{\textbf{0.0145}} & \textcolor{green}{\textbf{0.0071}} & \textcolor{green}{\textbf{0.0145}} & \textcolor{green}{\textbf{0.0365}} & \textcolor{green}{\textbf{0.0202}} & 0.0932 \\
\textbf{G3} & \textcolor{green}{\textbf{0.0256}} & \textcolor{green}{\textbf{0.0176}} & \textcolor{green}{\textbf{0.0120}} & \textcolor{green}{\textbf{0.0176}} & 0.0879 & \textcolor{green}{\textbf{0.0256}} & 0.0795 \\
\textbf{G4} & \textcolor{green}{\textbf{0.0260}} & \textcolor{green}{\textbf{0.0260}} & \textcolor{green}{\textbf{0.0152}} & \textcolor{green}{\textbf{0.0260}} & 0.0867 & \textcolor{green}{\textbf{0.0260}} & 0.1061 \\
\textbf{G5} & 0.2619 & 0.1667 & 0.1667 & 0.1667 & 0.7822 & 0.2619 & \textcolor{green}{\textbf{0.0238}} \\
\textbf{G6} & 0.1429 & 0.2857 & 0.1429 & 0.2857 & \textcolor{green}{\textbf{0.0079}} & 0.6429 & 0.0714 \\
\bottomrule
\end{tabular}
\end{table*}


\begin{table*}[!htb]
\centering
\caption{Separability results obtained for groups G1--G6 in the 1.5--2 kHz frequency band with the proposed method. $p$-values are obtained from hypothesis testing over spectral features extracted from XAI-driven \textit{Weighted Spectrograms}. Statistical significance is considered for $p < 0.05$ and highlighted using green, boldfaced font.}
\label{tab:features_band4}
\begin{tabular}{lccccccc}
\toprule
& \multicolumn{7}{c}{\textbf{Frequency Band 1.5--2 kHz}} \\
\midrule
& \textbf{$RP_4$} & \textbf{$SpBW_4$} & \textbf{$SpCF_4$} & \textbf{$SpF_4$} & \textbf{$SpFx_4$} & \textbf{$SpRE_4$} & \textbf{$SpR_4$} \\
& Th=0.7 & Th=0.8 & Th=0.8 & Th=0.8 & Th=0.9 & Th=0.8 & Th=0.6 \\
\midrule
\textbf{G1} & \textcolor{green}{\textbf{0.0071}} & 0.1490 & 0.1215 & 0.1215 & 0.6605 & \textcolor{green}{\textbf{0.0202}} & 0.0913 \\
\textbf{G2} & \textcolor{green}{\textbf{0.0019}} & \textcolor{green}{\textbf{0.0048}} & \textcolor{green}{\textbf{0.0048}} & \textcolor{green}{\textbf{0.0273}} & \textcolor{green}{\textbf{0.0365}} & \textcolor{green}{\textbf{0.0145}} & \textcolor{green}{\textbf{0.0045}} \\
\textbf{G3} & \textcolor{green}{\textbf{0.0048}} & \textcolor{green}{\textbf{0.0120}} & \textcolor{green}{\textbf{0.0048}} & \textcolor{green}{\textbf{0.0256}} & \textcolor{green}{\textbf{0.0496}} & \textcolor{green}{\textbf{0.0176}} & \textcolor{green}{\textbf{0.0028}} \\
\textbf{G4} & \textcolor{green}{\textbf{0.0043}} & \textcolor{green}{\textbf{0.0152}} & \textcolor{green}{\textbf{0.0087}} & \textcolor{green}{\textbf{0.0260}} & \textcolor{green}{\textbf{0.0459}} & \textcolor{green}{\textbf{0.0260}} & \textcolor{green}{\textbf{0.0152}} \\
\textbf{G5} & 0.1667 & 0.1667 & 0.0952 & 0.2619 & 0.0962 & 0.1667 & \textcolor{green}{\textbf{0.0238}} \\
\textbf{G6} & 0.0714 & 0.0714 & 0.2857 & 0.4286 & 0.1125 & 0.2857 & 0.0714 \\
\bottomrule
\end{tabular}
\end{table*}

\begin{table*}[!htb]
\centering
\caption{Separability results obtained for groups G1--G6 in the 2--4.41 kHz frequency band with the proposed method. $p$-values are obtained from hypothesis testing over spectral features extracted from XAI-driven \textit{Weighted Spectrograms}. Statistical significance is considered for $p < 0.05$ and highlighted using green, boldfaced font.}
\label{tab:features_band5}

\begin{tabular}{lccccccc}
\toprule
& \multicolumn{7}{c}{\textbf{Frequency Band 2--4.41 kHz}} \\
\midrule
& \textbf{$RP_5$} & \textbf{$SpBW_5$} & \textbf{$SpCF_5$} & \textbf{$SpF_5$} & \textbf{$SpFx_5$} & \textbf{$SpRE_5$} & \textbf{$SpR_5$} \\
& Th=0.7 & Th=0.7 & Th=0.7 & Th=0.7 & Th=0.9 & Th=0.7 & Th=0.7 \\
\midrule
\textbf{G1} & 0.0782 & \textcolor{green}{\textbf{0.0273}} & 0.0782 & 0.0616 & 0.3011 & \textcolor{green}{\textbf{0.0145}} & \textcolor{green}{\textbf{0.0273}} \\
\textbf{G2} & \textcolor{green}{\textbf{0.0071}} & 0.1490 & \textcolor{green}{\textbf{0.0145}} & 0.0616 & \textcolor{green}{\textbf{0.0365}} & 0.0983 & \textcolor{green}{\textbf{0.0202}} \\
\textbf{G3} & \textcolor{green}{\textbf{0.0120}} & 0.0879 & \textcolor{green}{\textbf{0.0176}} & 0.0663 & 0.0663 & \textcolor{green}{\textbf{0.0496}} & \textcolor{green}{\textbf{0.0176}} \\
\textbf{G4} & \textcolor{green}{\textbf{0.0152}} & \textcolor{green}{\textbf{0.0260}} & \textcolor{green}{\textbf{0.0411}} & 0.0649 & 0.0604 & \textcolor{green}{\textbf{0.0152}} & \textcolor{green}{\textbf{0.0152}} \\
\textbf{G5} & 0.1667 & 0.9048 & 0.0952 & 0.3840 & 0.3669 & 0.7143 & 0.2619 \\
\textbf{G6} & 0.1429 & 1.000 & 0.2857 & 0.4286 & 0.1772 & 1.000 & 0.4286 \\
\bottomrule
\end{tabular}
\end{table*}

\begin{table*}[!htb]
\centering
\caption{Separability results obtained for groups G1--G6 in the overall 0--4.41 kHz frequency band~\cite{amado2025}. $p$-values are obtained from hypothesis testing over spectral features extracted from XAI-driven \textit{Weighted Spectrograms}. Statistical significance is considered for $p < 0.05$ and highlighted using green, boldfaced font. Thresholds used for each significant feature are: $RP_B$=0.7, $SpBW_B$=0.7, $SpCF_B$=0.7 (0.5 for G1), $SpF_B$=0.6, $SpFx_B$=0.9 (0.5 for G5,G6), $SpRE_B$=0.7 (0.5 for G5) and $SpR_B$=0.7.}
\label{tab:bandaGlobal}
\begin{tabular}{lccccccc}
\toprule
& \multicolumn{7}{c}{\textbf{Frequency Band 0-4.41 kHz}~\cite{amado2025}} \\
\midrule
& \textbf{$RP_B$} & \textbf{$SpBW_B$} & \textbf{$SpCF_B$} & \textbf{$SpF_B$} & \textbf{$SpFx_B$} & \textbf{$SpRE_B$} & \textbf{$SpR_B$} \\
\midrule
\textbf{G1} & 0.1088 & 0.0702 & \textcolor{green}{\textbf{0.0330}} & 0.0553 & 0.4747 & 0.0702 & 0.0878 \\
\textbf{G2} & \textcolor{green}{\textbf{0.0273}} & \textcolor{green}{\textbf{0.0202}} & \textcolor{green}{\textbf{0.0103}} & \textcolor{green}{\textbf{0.0011}} & 0.3011 & \textcolor{green}{\textbf{0.0145}} & \textcolor{green}{\textbf{0.0103}} \\
\textbf{G3} & \textcolor{green}{\textbf{0.0360}} & \textcolor{green}{\textbf{0.0256}} & \textcolor{green}{\textbf{0.0256}} & \textcolor{green}{\textbf{0.0028}} & 0.2721 & \textcolor{green}{\textbf{0.0176}} & \textcolor{green}{\textbf{0.0120}} \\
\textbf{G4} & \textcolor{green}{\textbf{0.0260}} & \textcolor{green}{\textbf{0.0152}} & \textcolor{green}{\textbf{0.0087}} & \textcolor{green}{\textbf{0.0152}} & 0.3095 & \textcolor{green}{\textbf{0.0152}} & \textcolor{green}{\textbf{0.0152}} \\
\textbf{G5} & 0.3810 & 0.3818 & 0.2619 & \textcolor{green}{\textbf{0.0238}} & 0.2619 & \textcolor{green}{\textbf{0.0238}} & \textcolor{green}{\textbf{0.0238}} \\
\textbf{G6} & 0.2857 & 0.2857 & 0.0714 & 0.0714 & 0.4286 & 0.0714 & 0.0714 \\
\bottomrule
\end{tabular}
\end{table*}

\section{Discussion}
\label{sec:Disc}

In this section, we present and discuss the results, comparing them feature by feature with the closest approach in~\cite{amado2025} (subsections~\ref{sec:RP} -- ~\ref{sec:SPR}), followed by an overall summary in subsection~\ref{sec:conc}. Additional related work is reviewed in subsection~\ref{sec:related}, and final concluding remarks are provided in subsection~\ref{sec:conc2}.

\subsection{Relative Power}
\label{sec:RP}

\begin{figure*}[htb!]
\centering
    \includegraphics[width=0.9\textwidth]{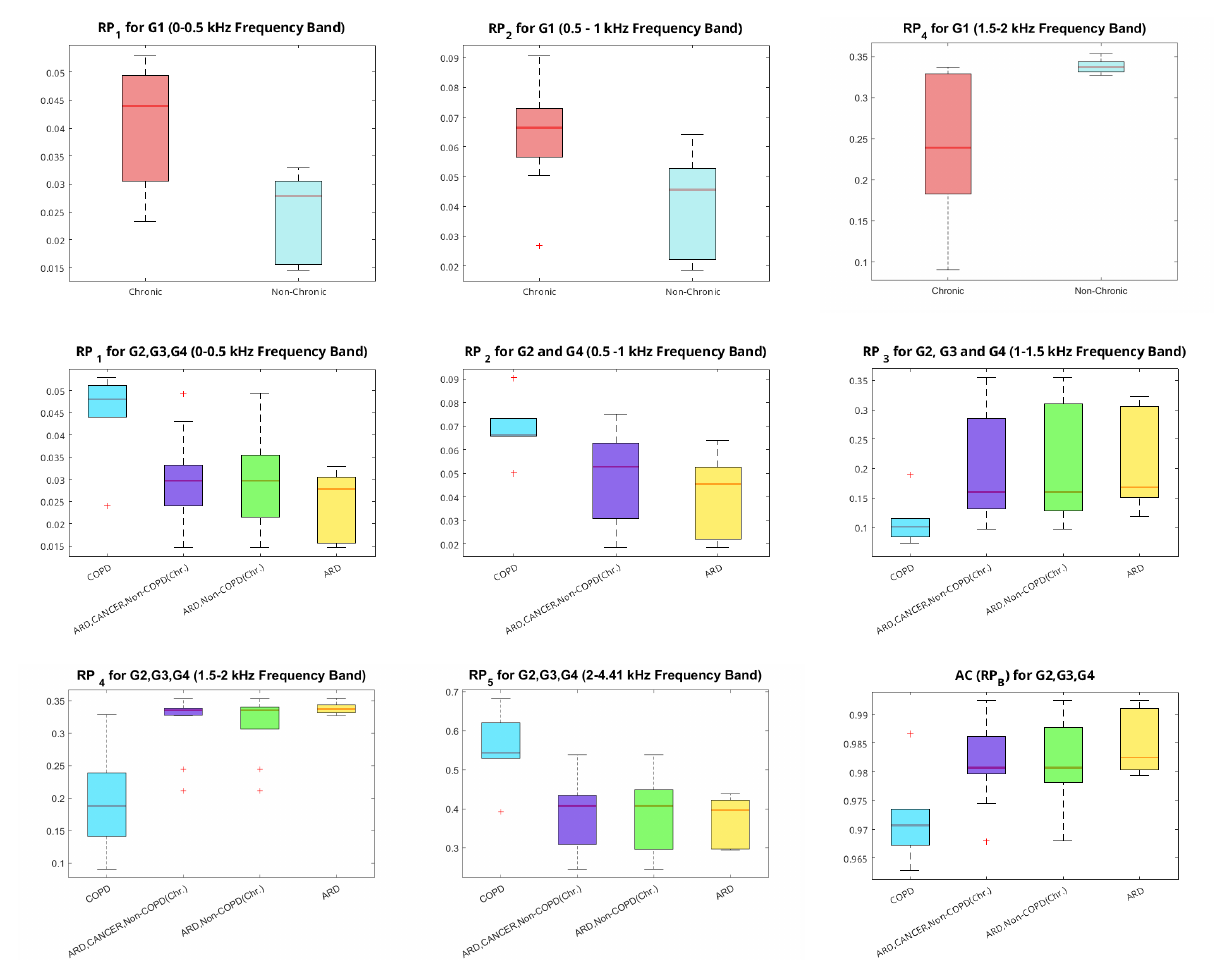}
  \caption{Boxplots obtained for Relative Power ($RP_j$) for significant features $p<0.05$.}
\label{fig:RPBox_1}
\end{figure*}

As shown in Figure~\ref{fig:RPBox_1} and summarized in Table \ref{tab:features_resume1} and \ref{tab:features_resume2}, the Relative Power boxplots for group G1 indicate that chronic patients exhibit significantly higher power in the low-frequency bands 1 and 2 compared to non-chronic patients, while power is lower in the 4th frequency band. Additionally, COPD patients display higher relative power in bands 1, 2, and 5 compared to other diseases (G2), other diseases excluding cancer (G3; except for band 2), and ARD/pneumonia (G4). In contrast, the opposite pattern is observed in bands 3 and 4, where relative power is lower for patients with COPD. As for the whole $[0-4.41)$ kHz band, $RP_B$ can be observed to be significantly lower for COPD patients in groups G2--G4. These consistent patterns illustrated in table \ref{tab:features_resume2}, mean that bands 3 and 4 have a more significant impact in the overall AC power differences between COPD and other diseases, driving $RP_B$ to lower values in the COPD group. Interestingly, the larger differences observed in the 5th band compared to the overall AC power suggest greater variability in the XAI-weighted spectrum for patients with COPD. Although lower overall AC power might imply the opposite, higher values of $RP_5$ in the 2–4.41 kHz range indicate substantial energy content at the upper end of the spectrum, contributing to this increased variability. This effect appears to be balanced by higher relative power values in the lower bands (1st and 2nd) and lower values in the middle bands (3rd and 4th), as we can observe in table \ref{tab:features_resume1}.

\subsection{Spectral Bandwidth}

\begin{figure*}[htb!]
\centering
        \includegraphics[width=0.9\textwidth]{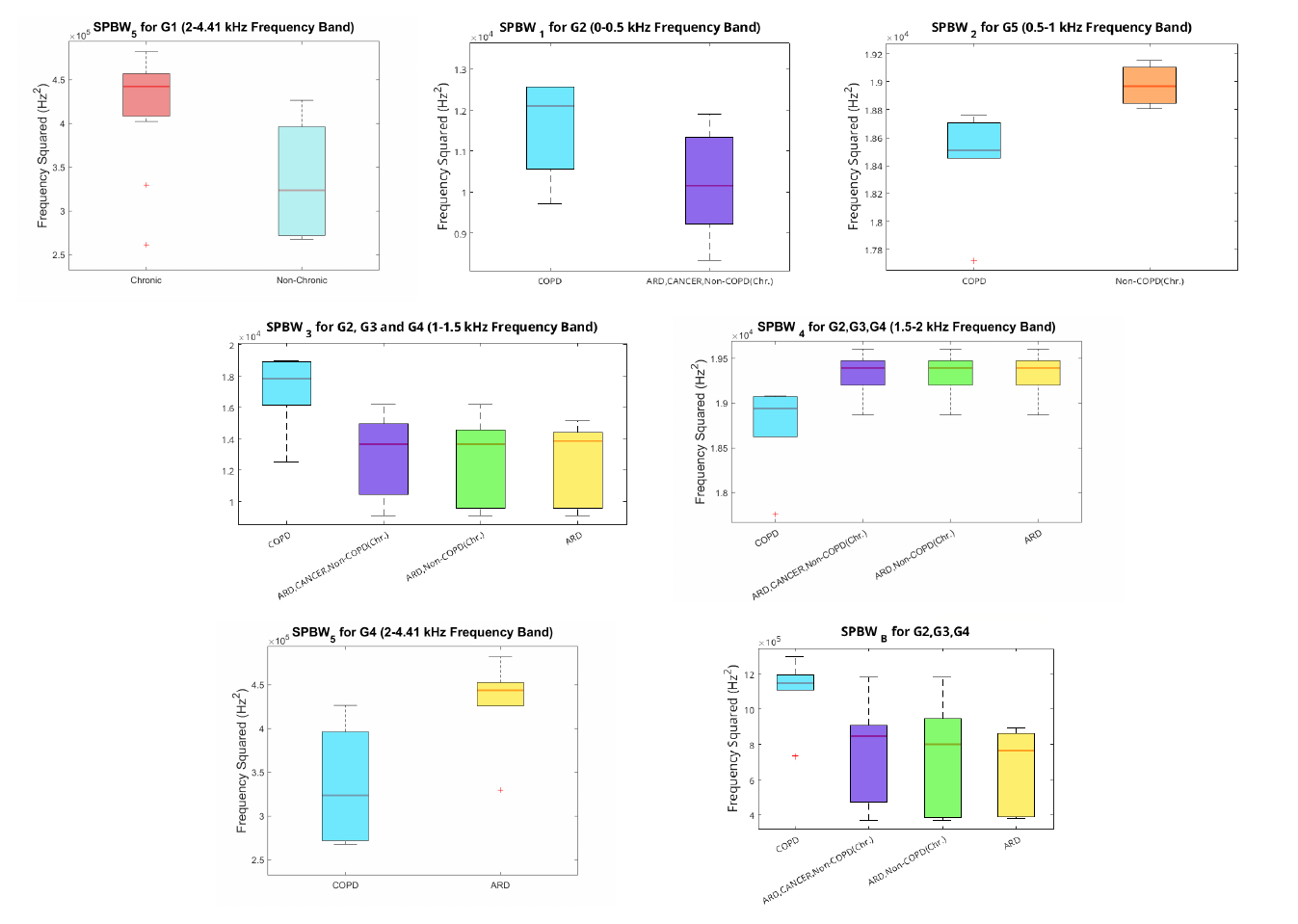}
  \caption{Boxplots obtained for Spectral Bandwidth ($SpBW_j$) for significant features $p<0.05$.}
\label{fig:SPBWBox_1}
\end{figure*}

As shown in Figure~\ref{fig:SPBWBox_1} and table \ref{tab:features_resume1}, for group G1, chronic patients exhibit a significantly higher spectral bandwidth (indicating more dispersed energy content) in the 5th band compared to non-chronic patients. . Patients with COPD show a significantly higher bandwidth in the 2nd band compared to other diseases (G2), and in the 3rd band compared to other diseases (G2), other diseases excluding cancer (G3), and ARD/pneumonia (G4). In contrast, in the next (4th) band, COPD patients present a significantly lower bandwidth when G2–G4 are studied. Lower bandwidth is also evident in bands 2 and 5 for G5 (COPD vs. other chronic diseases) and G4 (ARD), respectively. When considering the overall spectral band, COPD patients display a significantly higher bandwidth across G2–G4, with no overlap. These observations align with the $p$-values reported in Tables~\ref{tab:features_band2}--\ref{tab:bandaGlobal}. Apart from band 2, there is no overlap between the boxplots for patients with COPD and those of other groups.

\subsection{Spectral Crest Factor}

\begin{figure*}[htb!]
\centering
  \includegraphics[width=0.9\textwidth]{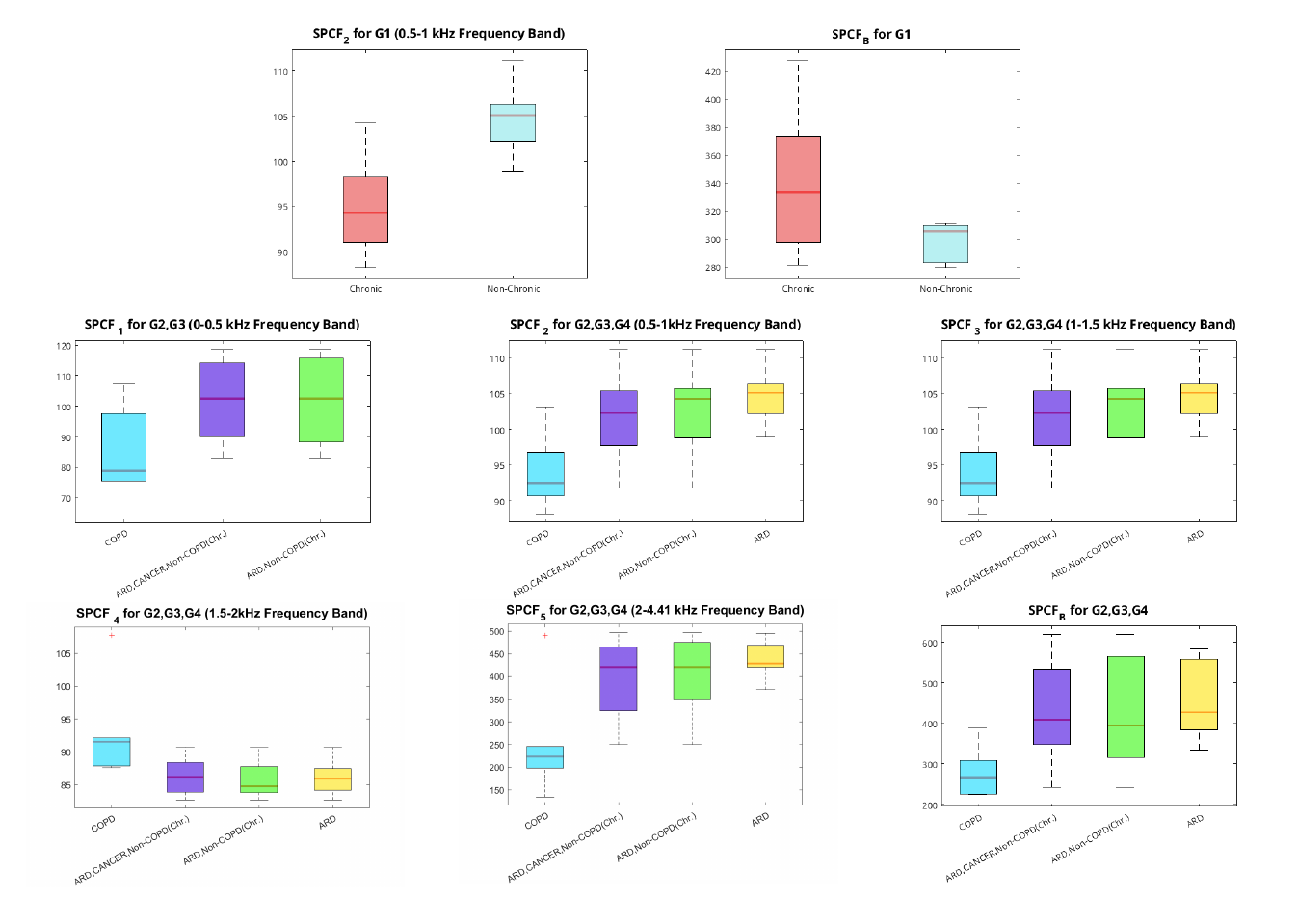}
  \caption{Boxplots obtained for Spectral Cress Factor ($SpCF_j$) for significant features $p<0.05$.}
\label{fig:SPCFBox_1}
\end{figure*}

The Spectral Crest Factor ($SpCF$) shows significantly lower values in the second frequency band for chronic patients compared to non-chronic patients in the G1 group comparison (see Figure~\ref{fig:SPCFBox_1}, top row, left-hand side). This indicates a higher concentration of energy at lower frequencies within this band. In contrast, the opposite trend is observed when considering the entire frequency range (top row, right-hand side), although boxplots reveal a considerable overlap in this case. 

For the remaining groups, significant results are also observed for G2–G4. Specifically, patients with COPD  exhibit lower $SpCF$ values than patients with other diseases (G2), other diseases excluding cancer (G3), and ARD/pneumonia (G4) in bands 2, 3, and 5, which drives the same trend for the overall frequency range. Significantly lower values are also observed in the first band, but only for groups G2 and G3, and with overlap in the boxplots. Conversely, significantly higher $SpCF$ values are found in band 4. All these findings align with the patterns documented in tables \ref{tab:features_resume1} and \ref{tab:features_resume2}.

\subsection{Spectral Flatness}

\begin{figure*}[htb!]
\centering
    \includegraphics[width=0.9\textwidth]{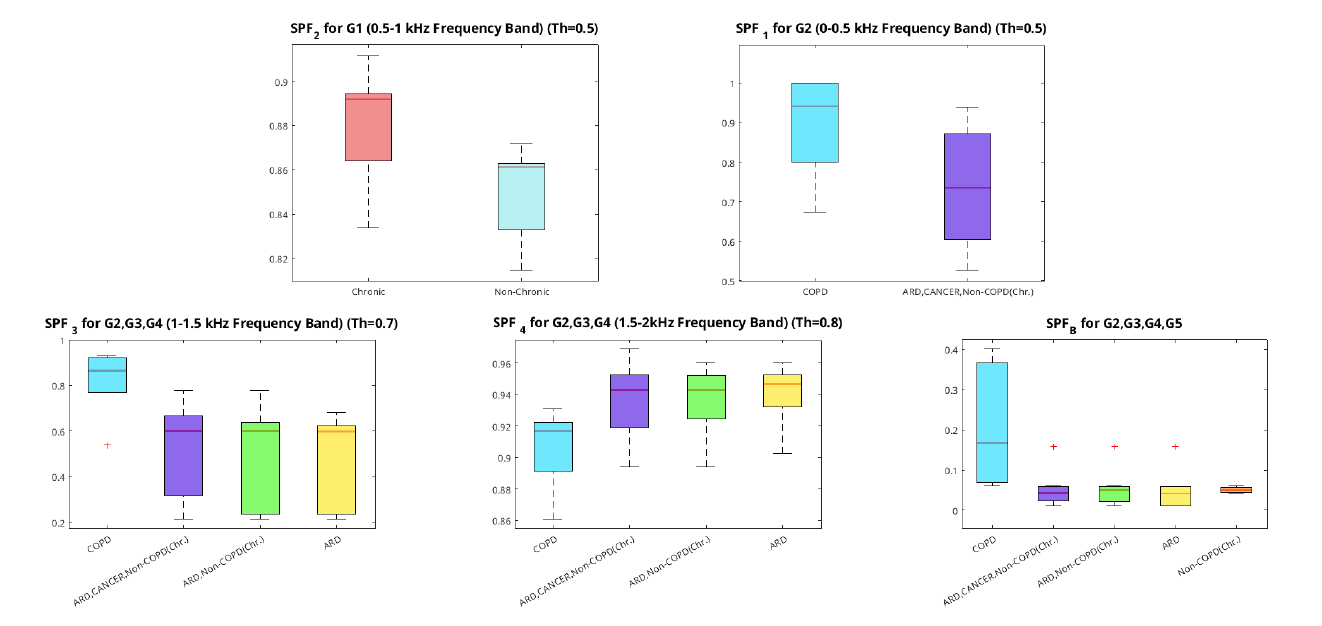}
  \caption{Boxplots obtained for Spectral Flatness ($SpF_j$) for significant features $p<0.05$.}
\label{fig:SPFBox_1}
\end{figure*}

In the case of Spectral Flatness ($SpF$), Figure~\ref{fig:SPFBox_1} displays a pattern similar to that observed for other features. Significant differences in the comparison group G1 are found for band 2, where chronic patients exhibit a flatter spectrum (more noise-like) than non-chronic patients. The lower frequency band also shows relevant differences between patients with COPD (higher $SpF$) and patients with other diseases (G2), although a significant overlap is evident in the boxplots. Bands 3 and 4 present meaningful differences for patients with COPD compared to subgroups in G2–G4; however, the trends are opposite: $SpF$ is higher (more noise-like) for  patients with COPD in band 3, but lower in band 4. A similar opposite behaviour is also observed in previously analysed features ($SpBW$ and $SpCF$), although the trends differ for the latter. Finally, when considering the overall frequency range, patients with COPD show significantly higher $SpF$ values than patients in the other subgroups in G2–G4 (presumably driven by the third band) and other chronic patients (study group G5).

\subsection{Spectral Flux}

\begin{figure*}[htb!]
\centering
   \includegraphics[width=0.9\textwidth]{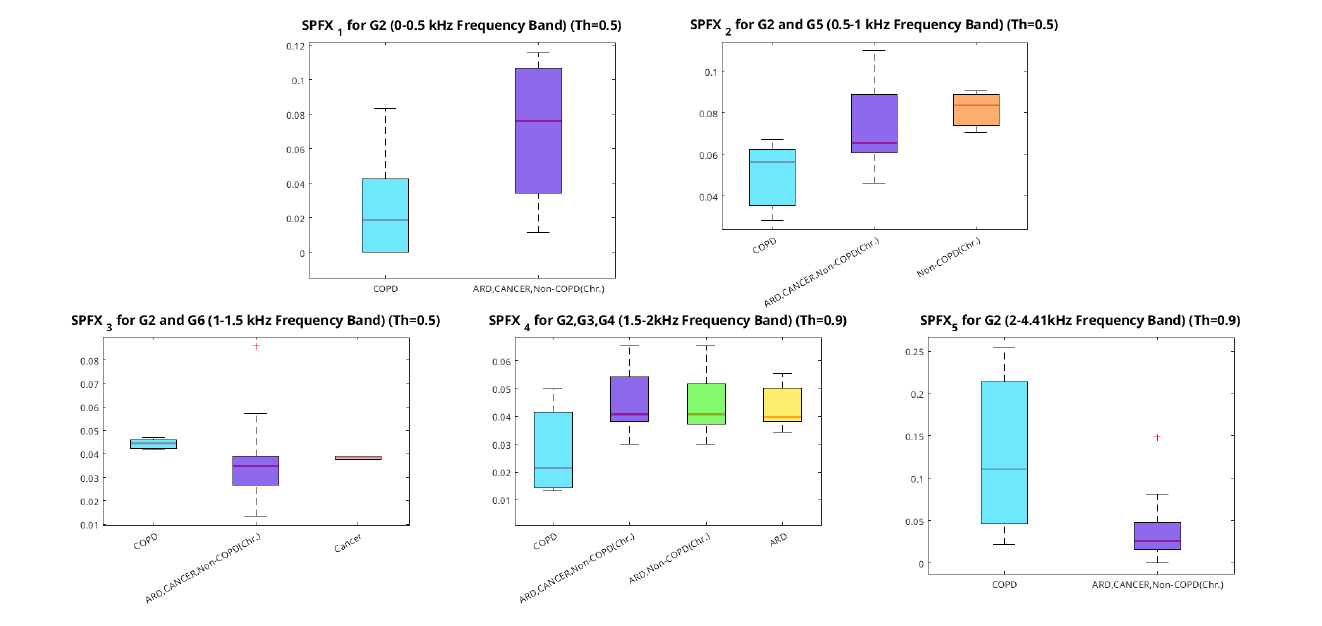}
  \caption{Boxplots obtained for Spectral Flux ($SpFX_j$) for significant features $p<0.05$.}
\label{fig:SPFXBox_1}
\end{figure*}

Spectral Flux was an uninformative feature in the global frequency analysis (see Table~\ref{tab:bandaGlobal}), showing considerable overlap between groups. This transformation is exemplified in tables \ref{tab:features_resume1} and \ref{tab:features_resume2}.
 However, the specific band analysis reveals clear differences, as shown in Figure~\ref{fig:SPFXBox_1}. Patients with COPD show significantly lower $SpFx$ values than patients with other diseases (G2) in bands 1, 2, and 4. Higher values of this feature represent greater temporal variation in the spectra. The observed pattern suggests that COPD is characterized by a spectral redistribution of temporal variability, where the frequency components in bands 1, 2, and 4 become more rigid compared to the less dominant ones under typical conditions. In contrast, the opposite trend is observed in bands 3 and 5. These findings demonstrate that sub-band analysis can uncover spatio-temporal patterns that are not evident in the overall frequency band, possibly due to the compensation of inter-band variability.

The lower $SpFx$ values for COPD patients are also significant for group G5 (non-COPD chronic patients) in band 2, and for the other subgroups in groups G3 (other diseases excluding cancer) and G4 (ARD/pneumonia) in band 4. Interestingly, Patients with COPD also exhibit significantly higher $SpFx$ values than cancer patients (comparison group G6) in band 3, although this result should be interpreted with caution due to the small size of the cancer group.

\subsection{Spectral Rényi Entropy}
\begin{figure*}[htb!]
\centering
   \includegraphics[width=0.9\textwidth]{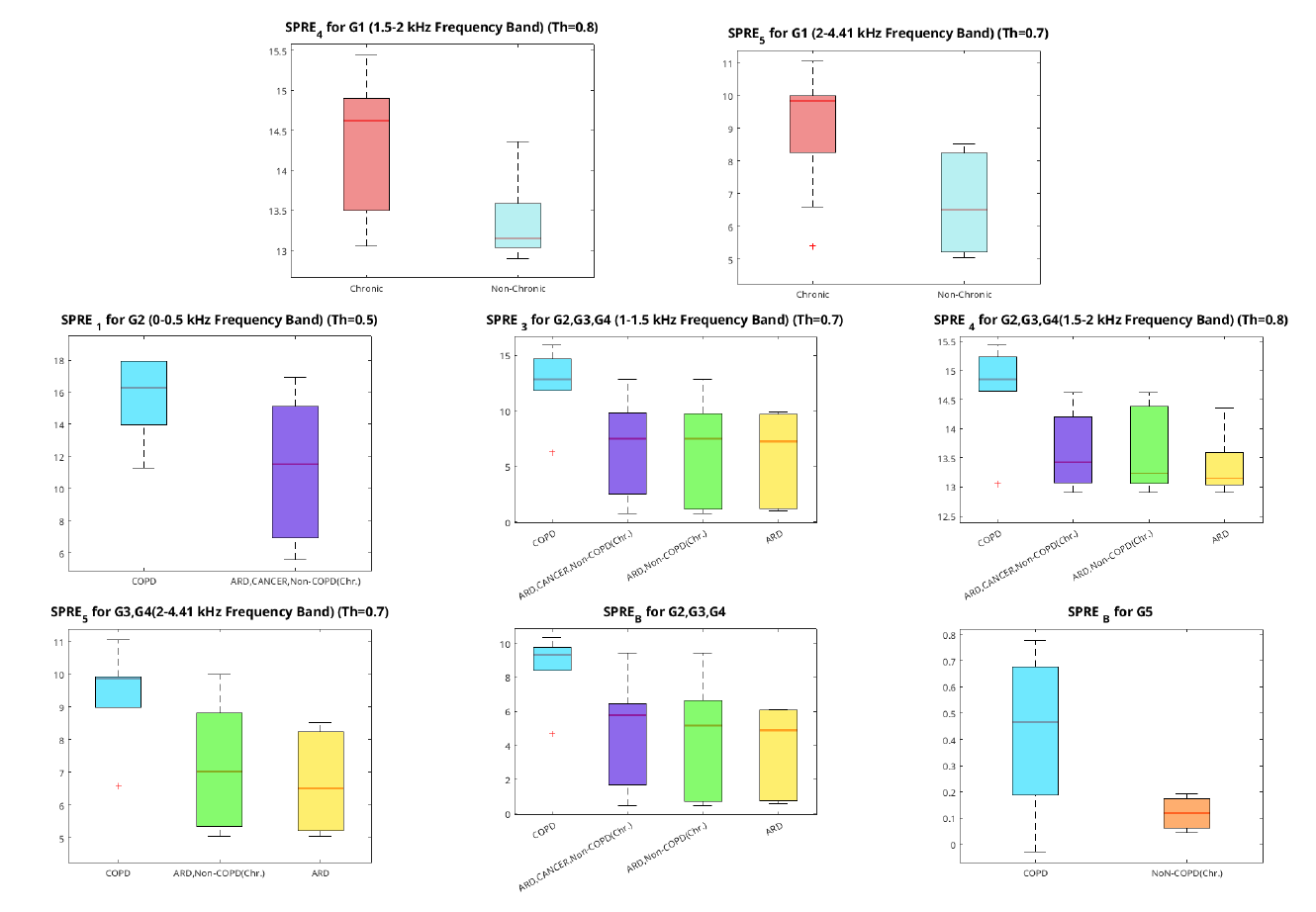}
  \caption{Boxplots obtained for Spectral Renyi Entropy ($SpRE_j$) for significant features $p<0.05$.}
\label{fig:SPREBox_1}
\end{figure*}

Spectral Rényi Entropy ($SpRE$) also proves to be a highly discriminative feature for distinguishing chronic diseases from non-chronic cases, as well as COPD from other disease subgroups across the different comparison groups, as evidenced by the non-overlapping boxplots in Figure~\ref{fig:SPREBox_1}. Chronic patients exhibit significantly higher $SpRE$ values than non-chronic patients in bands 4 and 5, indicating a more chaotic spectrum in those bands. A similar trend is observed for  patients with COPD, who show significantly higher $SpRE$ values than patients with other diseases (G2) in bands 1, 3, 4, and the overall frequency range; this trend extends to groups G3 (other diseases excluding cancer) and G4 (ARD/pneumonia) in bands 3, 4, 5, and the whole band. Notably, this feature consistently shows the same behaviour across all significant bands. Finally, $SpRE$ is also significantly higher for patients with COPD compared to other chronic patients (G5), when considering the entire frequency range. This exceptional discriminative power is reflect in \ref{tab:features_resume1}, where $SpRE$ shows significant group differences across all frequency bands.

\subsection{Spectral Roll-Off}
\label{sec:SPR}

\begin{figure*}[htb!]
\centering
   \includegraphics[width=0.9\textwidth]{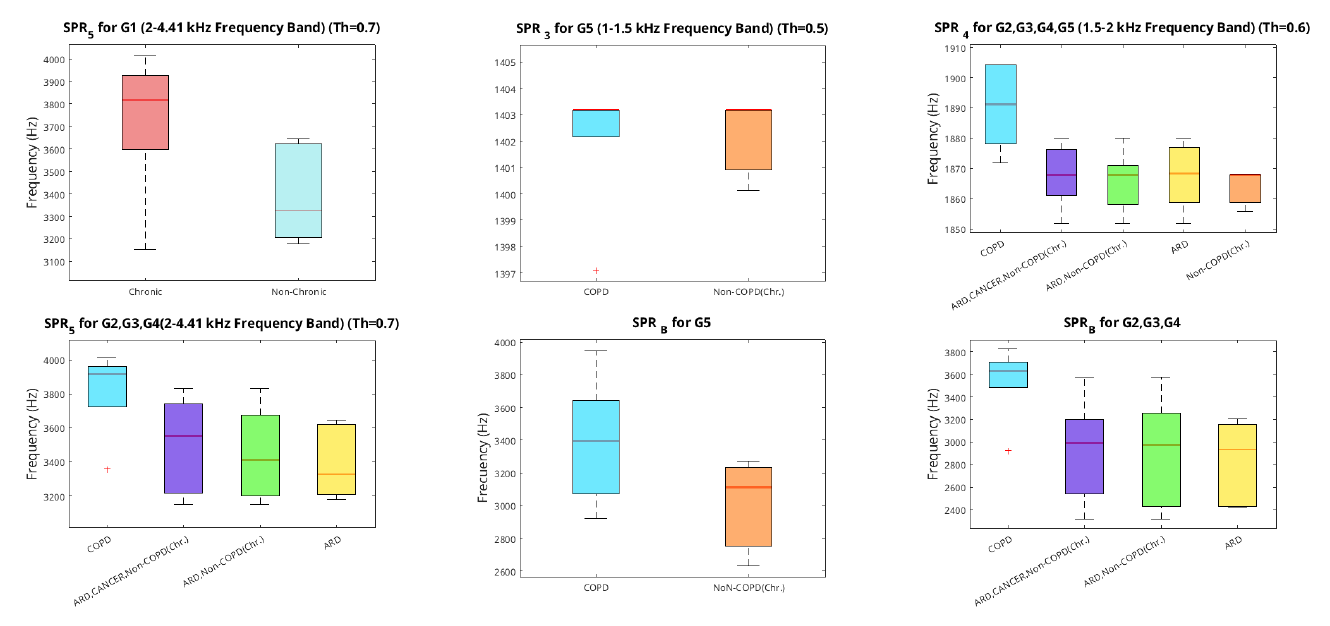}
  \caption{Boxplots obtained for Spectral Roll-Off ($SpR_j$) for significant features $p<0.05$.}
\label{fig:SPR_3}
\end{figure*}

Considering Spectral Roll-Off,the discriminative patterns summarized in table \ref{tab:features_resume1} can be validated observing the boxplots in Figure~\ref{fig:SPR_3}. This feature represents the frequency below which 85\% of the total power is accumulated. The boxplots show significantly higher $SpR$ values for chronic patients compared to non-chronic patients (G1) in band 5. Similarly, COPD patients show higher values than other chronic patients (G5) in bands 3, 4, and the overall frequency range, with no boxplot overlap in band 4. Comparisons of COPD patients with the subgroups in G2–G4 also reveal significantly higher $SpR$ values for COPD in bands 4, 5, and the whole frequency range. A higher $SpR$ indicates that the spectral energy is distributed towards higher frequencies rather than being concentrated at lower frequencies within the respective bands.

\subsection{Overall discussion}
\label{sec:conc}

The Spectral Features analysis over frequency bands reveals cough specifics associated to different pathologies. Looking at tables \ref{tab:features_resume1} and \ref{tab:features_resume2}  we can assess whether a feature is discriminative for each study group. For G1, $RP$ and $SpR$ are particularly interesting: although they are discriminative in two or more individual bands, they do not appear as discriminative in the global band. A similar pattern is observed for $SpFx$ in Group G2, where the feature is discriminative across all individual bands but not in the global analysis. This behavior may be explained by the compensation of frequency values across bands. Taking the case of $SpFx$, we observe that in bands 1, 2, and 4, C1 presents lower values than C2, while in the remaining bands, the opposite occurs. This opposing behavior across bands likely neutralizes the overall effect in the global analysis, thereby diminishing the feature’s discriminative power at the global.

Another key observation from a detailed analysis of the features is the influence of each band on the overall frequency pattern. For instance, in the case of $RP$ for groups G2, G3, and G4, although the bands might initially appear to counterbalance one another, the intermediate bands (band 3 and band 4) have a dominant effect, shaping the global frequency distribution. This leads to significantly lower COPD values (C1 in these groups), which are critical for the diagnosis of the pathology.

In the case of $SpBW$ and $SpCF$, an opposing trend is observed in band 4 for COPD; however, this does not significantly impact the overall analysis, as the general pattern is primarily defined by the other bands.

As a summary: 

\begin{itemize}
    \item Some features, such as $RP$ and $SpR$ in G1, and $SpFx$ in G2, are discriminative at the individual band level but not in the global analysis. This suggests that important local information may be lost when only considering the global band as in~\cite{amado2025}, probably due to opposing trends in different frequency bands.
    \item Individual frequency bands (especially intermediate ones like band 3 and band 4) may have a dominant influence on the global frequency pattern. This influence may lead to significant variations in key metrics which are crucial for identifying pathological patterns.
\end{itemize}

\subsection{Related work}
\label{sec:related}

In~\cite{wullenweber2022coughlime}, audio-cough spectrograms were decomposed into interpretable representations, which were further divided into equal intervals. Each interval was analyzed for its importance in diagnosing COVID-19 using a deep ResNet model. The analysis was performed by randomly flipping interval information to measure how positional changes affected prediction accuracy. Four decompositions were considered: loudness, temporal, spectral, temporal-loudness, and Non-Negative Matrix Factorization. The results showed that changes in position of temporal information had the highest impact, supporting our hypothesis that spectral features offer a more robust alternative by reducing dependence on precise cough onset and offset detection.

The work in~\cite{Shen2024} employed Grad-CAM to visualize the contribution of CNN-extracted features when distinguishing COVID-19, asthma, pneumonia, and healthy subjects. The visualizations suggested possible explanations for model decisions, showing that high-frequency features were influential for COVID-19 coughs, medium-frequency features for healthy coughs, and low-frequency features for pneumonia coughs. While informative, this approach focused only on identifying which spectral regions were most relevant for each disease class, without explaining the underlying nature of the features, since they were directly provided by the deep learning model. In contrast, our work identifies physically meaningful spectral features in each frequency band, offering deeper insights into the characteristics of COPD and other chronic conditions.

Similarly, ~\cite{avila2021investigating} used CAM as a post-hoc explanation method, highlighting the most influential regions in input log-Mel spectrograms. The results showed that the model primarily focused on the actual cough segments rather than extraneous factors such as noise or zero-padding. Score-CAM in particular indicated that the CNN concentrated on the second phase of cough events, consistent with prior research showing that key differences among cough types often emerge in this phase. However, this reliance on temporal segmentation makes CAM-based methods sensitive to cough onset detection, limiting robustness compared to spectral approaches as ours. Frequency-wise, this work reported that COVID-positive coughs were associated with low-frequency regions, while non-COVID samples emphasized higher frequencies. These findings contradict those of ~\cite{Shen2024}, illustrating that CAM-based methods can be model-dependent since they rely on gradients and convolutional activations. Our approach, based on occlusion maps, avoids these biases as it is model-agnostic, requiring only perturbations of the input and observations of the output to build the maps.

In ~\cite{sobahi2022explainable}, Grad-CAM was applied to visualize regions of Fractal Dimension (FD) images influencing a ViT model for classifying COVID-19, healthy, and symptomatic coughs. For COVID-19 samples, the model mainly focused on the Castiglioni and Higuchi FD regions; for healthy samples, the Castiglioni FD region; and for symptomatic cases, the Katz and Castiglioni FD regions. While effective in highlighting influential areas, the results were harder to interpret than the spectral features here studied, since FD images lack direct physical meaning.

The most comparable works to ours are ~\cite{amado2024,amado2025}, both of which employed occlusion map-weighted spectrograms to extract disease-discriminant features. In ~\cite{amado2024}, Gaussian Mixture Models (GMMs) were fitted to weighted spectrograms, and temporal parameters such as the mean $\eta_x$ and standard deviation $\sigma_x$ of the non-dominant Gaussian showed significant discriminative power between COPD and other disease groups ($p=0.0238$ and $p=0.0139$, respectively). However, because these features were temporal in nature, the method was highly sensitive to the precise onset and offset of cough events, limiting robustness. This limitation was addressed in ~\cite{amado2025}, which computed spectral features across the full frequency band $B\equiv[0, 4.41)$ kHz. These features not only provided greater separability for COPD patients but also revealed differences between chronic and non-chronic conditions not captured in \cite{amado2024}. Given these advantages, we adopted \cite{amado2025} as the benchmark for this study, reproducing its results for the sake of completeness (see Table ~\ref{tab:bandaGlobal} and Figures ~\ref{fig:RPBox_1}--\ref{fig:SPR_3}). Our results show that while global spectral analysis identifies general trends, frequency sub-band analysis yields additional insights, uncovering patterns that remain hidden at the full-band level (see subsections~\ref{sec:RP} -- ~\ref{sec:SPR}).

\subsection{Concluding Remarks}
\label{sec:conc2}

The combination of CNNs and XAI with feature-based spectral analysis in five separate frequency bands enhances the understanding of cough sound characteristics in the context of chronic respiratory diseases in general, and COPD in particular. Results show that patterns observed in the overall frequency band often follow different trends when analysed within specific bands. Thus, analyzing each band individually appears to yield insights that go beyond those obtained from global spectral analysis. Since our findings are derived from a limited dataset and show potential, further validation through studies with more extensive data is needed.

\section*{CRediT authorship contribution statement}
\textbf{P.A.C.}: Methodology, Data Curation, Formal Analysis, Writing -- Original Draft. 
\textbf{L.S.R.}: Investigation, Methodology, Writing -- Original Draft, Writing -- Review \& Editing.
\textbf{X.W.}: Writing -- Review \& Editing, Funding Acquisition. 
\textbf{J.G.L.}: Investigation, Resources, Validation, Funding Acquisition. 
\textbf{C.A.L.}: Methodology, Supervision, Writing -- Review \& Editing. 
\textbf{P.C.H.}: Conceptualization, Writing -- Review \& Editing, Supervision, Funding Acquisition.

\section*{Ethics statement}
The study was carried out in accordance to the Declaration of Helsinki and was approved by the Área de Salud de Palencia Research Ethics Committee meetings 17/05/2018 and 09/08/2023. Subjects provided their informed consent before the recordings.

\section*{Declaration of competing interest}
The authors declare no conflict of interest.

\section*{Funding}
This work was supported by projects TED2021-131536B-I00, PID2022-142166NA-I0, and CPP2021-008880, funded by the Spanish MCIN/AEI/10.13039/501100011033, with TED2021-131536B-I00 and CPP2021-008880 co-funded by the EU NextGenerationEU/PRTR. The work was also partially funded by GRS 2837/C/2023 funded by Gerencia Regional de Salud, Junta de Castilla y León, Spain, EU Horizon 2020 Research and Innovation Programme under the Marie Sklodowska-Curie grant agreement No. 101008297. This article reflects only the authors’ view. The European Union Commission is not responsible for any use that may be made of the information it contains.

\bibliographystyle{elsarticle-num} 
\bibliography{biblio_abbreviated}

\end{document}